\documentclass[letterpaper, 10 pt, conference]{ieeeconf}
\IEEEoverridecommandlockouts
\usepackage[utf8]{inputenc}
\usepackage[english]{babel}
\usepackage{amsmath}
\usepackage{amssymb}

\usepackage{graphicx}

\usepackage{multirow, multicol, makecell, booktabs}

\usepackage{arydshln}
\usepackage{hyperref}
\usepackage{siunitx}

\usepackage{color}

\newcommand{\ra}[1]{\renewcommand{\arraystretch}{#1}}
\newcommand{\cs}[1]{\renewcommand{\tabcolsep}{#1pt}}

\newcommand{\materialurl}{ \url{https://romarcg.github.io/traversability_estimation} }

\title{\LARGE \bf
Learning Ground Traversability from Simulations
}

\author{R. Omar Chavez-Garcia, J\'{e}r\^{o}me Guzzi, Luca M. Gambardella and Alessandro Giusti
\thanks{This research was supported by the Swiss National Science Foundation through the National Centre of Competence in Research (NCCR) Robotics.}
\thanks{R. Omar Chavez-Garcia, J\'{e}r\^{o}me Guzzi, Luca M. Gambardella and Alessandro Giusti are with the Dalle Molle Institute for Artificial Intelligence (IDSIA), USI-SUPSI, Lugano, Switzerland. {\tt\small omar@idsia.ch}}
\thanks{Data and code to reproduce our results, video demonstrations, media material, and additional information are available online: \materialurl}
}

\begin{document}

\bstctlcite{IEEEexample:BSTcontrol}

\maketitle              
\thispagestyle{empty}
\pagestyle{empty}

\begin{abstract}
Mobile ground robots operating on unstructured terrain must predict which areas of the environment they are able to pass in order to plan feasible paths.
We address traversability estimation as a heightmap classification problem: we build a convolutional neural network that, given an image representing the heightmap of a terrain patch, predicts whether the robot will be able to traverse such patch from left to right.
The classifier is trained for a specific robot model (wheeled, tracked, legged, snake-like) using simulation data on procedurally generated training terrains; the trained classifier can be applied to unseen large heightmaps to yield oriented traversability maps, and then plan traversable paths. We extensively evaluate the approach in simulation on six real-world elevation datasets, and run a real-robot validation in one indoor and one outdoor environment.
\end{abstract}
%


\section{Introduction}
\label{sec:introduction}

In most indoor scenarios, mobile robots are equipped with a map of the environment, which is divided into traversable and non-traversable cells. A cell containing movable obstacles or walls is labeled as \emph{non-traversable}, while a cell with no obstacles is labeled as \emph{traversable}. Using this internal map, path planning can be solved using well-known algorithms~\cite{PierreSermanet:2008tc}.

In outdoor scenarios, creating a similar map of the terrain might be challenging. A cell might be traversable only in a specific direction and by a specific robot: a wheeled robot with limited power might be able to descend, but not ascend, a slope; a tall robot with a high center of mass might both ascend and descend the same slope, but then capsize when traversing it from side to side; a bicycle might hop on a side-walk, but only if it approaches it from an orthogonal angle. Moreover, it might be difficult to anticipate all the difficulties that a robot may encounter: a legged robot may stumble in a terrain with holes of a size comparable to its feet; vacuum cleaner robots might get stuck over power cords; a car with a low chassis may not pass over a speed bump.

We consider the problem of estimating where and in which directions a given 3D terrain is traversable by a specific ground robot, using a general approach based on machine learning that applies regardless of the robot’s locomotion method (wheeled, tracked, legged, snake-like), physical characteristics (size, motor torque), and low-level controller (anti-skid algorithms for wheeled robots, foothold selection and gait selection algorithms for legged robots).
We will define a given terrain patch as \emph{locally traversable} in a given direction if the robot, once placed in the center of the patch, can proceed in that direction for at least a short distance when driven by its low-level control algorithm.

When robot control parameters are known, traversability is only affected by the characteristics of the terrain around the robot's position. Therefore, for a given robot pose $X^\text{robot}$ with position $p$ and orientation $\theta$, we consider a heightmap patch centered in $p$ and rotated in such a way that the robot is pointing towards the right of the patch.
The patch is represented as a heightmap image whose pixels indicate height values.
We cast the patch traversability estimation as a binary classification problem (with classes \emph{traversable} vs \emph{non-traversable}), using such image as input, and solve it by training a convolutional neural network (CNN).

Training datasets are obtained by simulating the robot on many procedurally generated training terrains, which represent a variety of obstacles such as ramps, steps, bumps, holes and rugged areas; during such simulations, the robot is spawned in random positions and orientations, and instructed to proceed straight ahead; its progress is monitored and instances for traversable (where the robot successfully proceeds) and non-traversable (where the robot can't proceed) heightmap patches are continuously recorded.  Once the model is learned from such training data, it can be applied densely on any, possibly large, unseen heightmap.  Because accurate physical simulation is expensive, evaluating the classifier on many points and orientations of a test terrain is orders of magnitude faster than simulating the robot.

After reviewing relevant literature in Section \ref{sec:relatedwork}, we introduce in Section~\ref{sec:traversabilityestimation} the \textbf{main contribution} of this work: a complete framework for traversability estimation. This framework includes: i) an algorithm for procedural generation of training heightmaps, ii) a classifier for estimating traversability of heightmap patches, and iii) its application to path planning.
Experimental validation and results on real-world heightmaps are described in Section~\ref{sec:experimentalresults}.

\section{Related Work}
\label{sec:relatedwork}

Estimating terrain traversability is a fundamental capability for animals and mobile ground robots~\cite{ugur2010}; most organisms are known to use visual perception for this task, and most related works in robotics use on-board cameras or depth sensors. Then, traversability can be estimated from appearance cues, 3D geometry, or both~\cite{Papadakis:2013ev}.  In many cases, the link between such input and traversability is learned in a supervised fashion. We first categorize related works by their input data, then focus on different options to gather training information.

\textbf{Inputs}. When using appearance cues as input, one possible approach is to learn a classifier to directly classify traversable areas of the terrain in front of the robot~\cite{Hudjakov:2009cv,Shirkhodaie:2008da}.
A more common approach is to first use a classifier to segment the input image in a number of classes (e.g., paved, rocky, grass, obstacle), then assign to each segment a predetermined cost for traversal, which may be infinite for segments known to be impassable~\cite{Brooks:2007gf}.  In either case, some works use generic visual features~\cite{Khan:2011jc} (such as texture descriptors) with standard classifiers (such as support vector machines or random forests~\cite{bishop2006}), whereas others adopt deep learning techniques such as CNNs~\cite{PierreSermanet:2008tc,Langkvist:2016dy}, which operate on raw image data and learn meaningful problem-specific visual features.

Geometry-based approaches use local~\cite{delmerico2016active} or global sensory data to derive an elevation map of the terrain, which is a convenient spatial representation for ground robots~\cite{Bellone:2016bz}. Then, one option is to simulate a model of the robot on different areas of such elevation map: this allows one to explicitly test traversability, or, in a simpler setup, to evaluate the pose that the robot would assume when lying on each point of the elevation map~\cite{Lacroix:2002ig}. A more common approach evaluates each point of the elevation map by extracting simple local features (such as slope, roughness, step height), and then estimate traversability either through handcrafted rules, or using a learned classifier~\cite{karumanchi2010,belter2013,wermelinger2016}.

Our approach uses exclusively geometry information (while in this paper we work under the assumption that the heightmap is given in advance, nothing would change if the data was acquired by an on-board sensor).  Instead of relying on high-level features, we feed raw elevation data within each heightmap patch to a CNN; to the best of our knowledge, this is the first application of deep learning to heightmap data for traversability estimation. By using a CNN to extract problem-specific features, we learn which terrain patterns could cause locomotion problems to a specific robot, without prior assumptions: as we discussed in Section~\ref{sec:introduction}, such patterns may be complex, orientation-dependent and counterintuitive; experimental results in Section~\ref{sec:classificationresults} confirm that our approach performs better than a feature-based one.

\textbf{Source of training data}. Using a supervised learning approach exempts one from the need to manually define the link between the input data and traversability; on the other hand, it requires a set of labeled examples, whose size, quality and representativeness are key to the final performance: the strategy adopted to acquire such training data is an important (and sometimes, prevailing) component of all related literature~\cite{Papadakis:2013ev}.
Our approach exclusively relies on training data acquired from simulations on procedurally-generated terrains: this allows us to cheaply generate large datasets for data-hungry deep learning models.  This approach has not yet been attempted in the traversability literature, but training from simulations is a common strategy for learning manipulation or legged locomotion skills, especially when reinforcement learning techniques are adopted~\cite{nips2017-learn2run,Peng2015}.

One possible drawback of this choice is that the procedurally generated terrains may not be representative of the shape of real terrains: this would yield classifiers that suffer from bad performance when predicting whether a simulated robot can traverse a testing terrain acquired from the real world; our experiments in Section~\ref{sec:classificationresults} refute this hypothesis. Another possible drawback is that the simulation (of the robot, the terrain, or their interaction) may not be realistic enough: then, a classifier that is accurate when predicting the traversability of a simulated robot would be inaccurate for a real robot. 
Our real robot experiments (Sections~\ref{sec:classificationresults} and~\ref{sec:pathplanning}) suggest that, in two different environments, the classifier outputs match the observed robot ability to traverse obstacles; however, such environments have the same easy characteristics of our simulated environment: non-slippery, solid terrain. The predictive performance of our current classifier on sandy, unstable or slippery environments would be poor, unless such characteristics are simulated during training.\footnote{This assumes that the whole terrain, both in training and in testing, has the same physical characteristics; if it does not, additional input would be needed, as we foresee in Section~\ref{sec:discussion}} Accurately predicting vehicles on soft ground is an important research line~\cite{Stentz:1998fv}, which may leverage ad-hoc simulators validated with real-world data~\cite{Huntsberger:uw}.

Transferring models learned in simulation to the real world is a very active research topic in the manipulation literature~\cite{andrei2017}. Instead of dealing with this issue, the traversability literature acquires training data from the real world; then, associates observed input data to a ground-truth traversability label, which can be determined in one of two ways.

The first option is to use the robot’s own experience.  Then, the robot needs a way to detect whether the area it is passing is traversable or not, using wheel slippage~\cite{Le:du1997}, vibration sensors~\cite{Bekhti:2014cn}, or visual odometry to check progress (which is a possible extension for our approach, as we note in Section~\ref{sec:discussion}).

A second option is to entrust labeling to a human; this could be done in a direct, straightforward way (the robot acquires data, a human marks each input with a traversability label, yielding a training set)~\cite{Giusti:2016ie}, or using strategies to make the process more efficient.  For example, in~\cite{Silver:2010cd} a human draws a path from a source to a target, and the system infers which areas the human purposefully avoided and automatically uses such patches as non-traversable examples.

An early version of the present work, using a simplified model and with a preliminary experimental evaluation, was previously presented in~\cite{Chavez-Garcia2017}.

\section{Traversability Estimation Framework}
\label{sec:traversabilityestimation}

\begin{figure}
  \centering
  \includegraphics[width=0.99\linewidth]{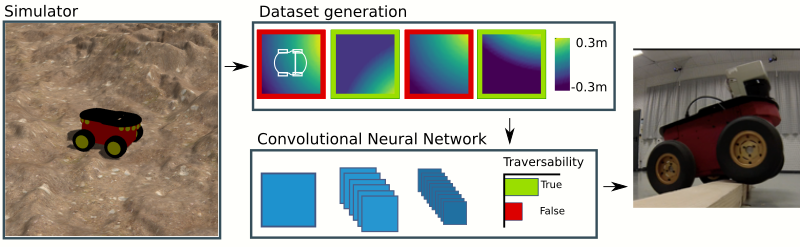}
  \caption{The robot model runs in simulation on procedurally generated terrains (left) to generate datasets linking heightmap patches with their traversability (top); on these datasets we train classifiers to estimate the probability that a given heightmap patch is traversable or not (bottom).
  The learned classifier correctly predicts terrain traversability for a real robot (right).}
  \label{fig:pipeline}
\end{figure}

Figure \ref{fig:pipeline} illustrates the proposed framework for traversability estimation. Next, we describe how the simulation environment is set-up (Section~\ref{sec:traversabilitysimulation}), detail the process for generating training and evaluation datasets (Section~\ref{sec:datasetgeneration}), and finally describe classification approaches (Section~\ref{sec:traversabilitylearning}).

\subsection{Traversability from Simulation}
\label{sec:traversabilitysimulation}

We adopt the Gazebo simulator and the ODE physics engine~\cite{gazebo2004} for accurate physical simulation.
We simulate the Pioneer 3-AT mobile robot (of size  $\SI{49}{cm}\times\SI{50}{cm}\times\SI{29}{cm}$, see Fig.~\ref{fig:pipeline}) moving forward at a constant velocity of $\SI{15}{\cm\per{s}}$ on an uneven terrain whose shape is determined by a given heightmap. We capture the robot's trajectory on the terrain to extract traversability information.

\subsubsection{Heightmap Generation}
\label{sec:heightmapgeneration}

In order to generate meaningful training data, we need to simulate the robot moving on interesting terrains that pose varied and representative challenges.  This could be achieved by using data from real terrains, by manually creating interesting heightmaps, and/or by synthesizing them using procedural generation techniques. We follow exclusively the third option.  This choice allows us to generate arbitrarily large amount of training data, and explore an interesting research question: can we learn a traversability classifier from synthetic heightmaps that works well on real heightmaps?

We generate 30 training terrains; the size of each generated heightmap is $\SI{512}{px}\times \SI{512}{px}$ that, when simulated, is scaled to represent a surface of $\SI{10}{m} \times \SI{10}{m}$ ($\approx2$ \si{\cm\per{px}} resolution).  Each terrain is generated by summing multiple realizations of random 2D simplex noise \cite{perlin2002improving} (a variant of Perlin noise \cite{lagae2010survey}
frequently adopted in the procedural generation literature \cite{smelik2009survey}) with different periods.  For example, a heightmap obtained from simplex noise with period \SI{30}{cm}, scaled such that it extends to a height of \SI{20}{cm}, yields a terrain with medium-sized smooth rocks; a period of \SI{10}{m} with a \SI{3}{m} height range yields a landscape with small steep hills. A weighted sum of the two components yields rocky hills.  Within a given terrain, we modulate the weight of one component such that it spans the range $0$ to $1$ along the $x$ axis, and the weight of the other component such that it does the same along the $y$ axis.  In this example, we would have flat terrain at $(x,y) = (0,0)$ (both components have zero weight), rocks on flat ground at $(x,y) = (10,0)$ (only the first component), smooth hills at $(x,y) = (0,10)$ (only the second component).  This ensures that a single map represents a range of parameters, including terrain that is neither too simple nor too challenging.

Additional features such as holes, steps, bumps protruding from a flat surface and rail-like indentations are generated by applying various scalar functions to heightmap values. Figure~\ref{fig:heightmapsexamples} shows examples of resulting training terrains.~\footnote{In supplementary material we provide source code for generating our training dataset and an appendix describing the approach in more detail.}

\begin{figure*}[t!]
  \centering
  \includegraphics[width=0.24\linewidth]{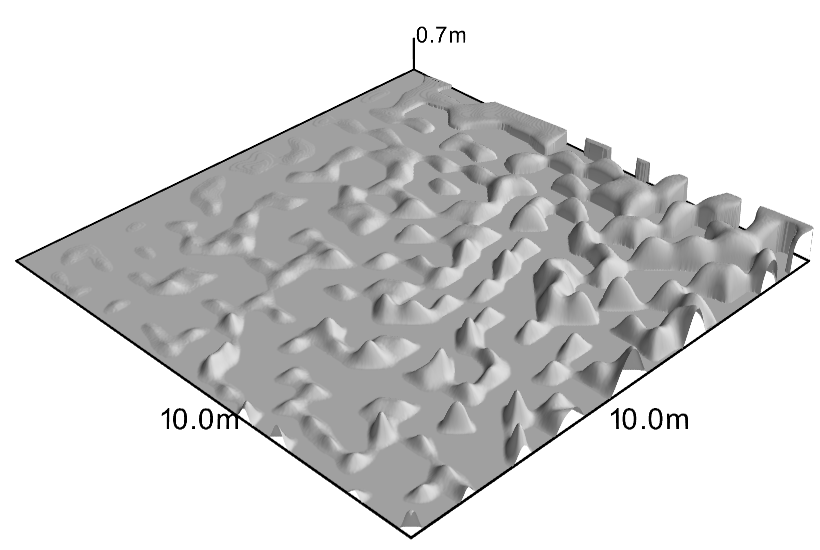}
  \includegraphics[width=0.24\linewidth]{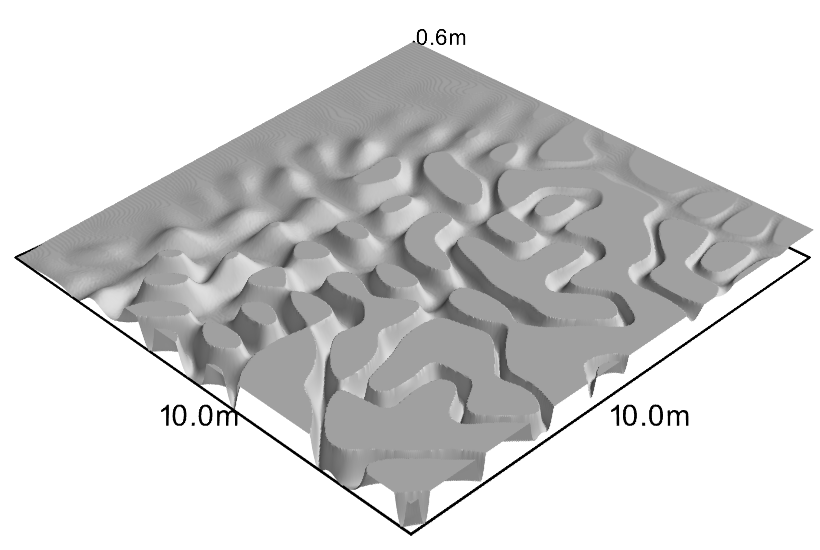}
  \includegraphics[width=0.24\linewidth]{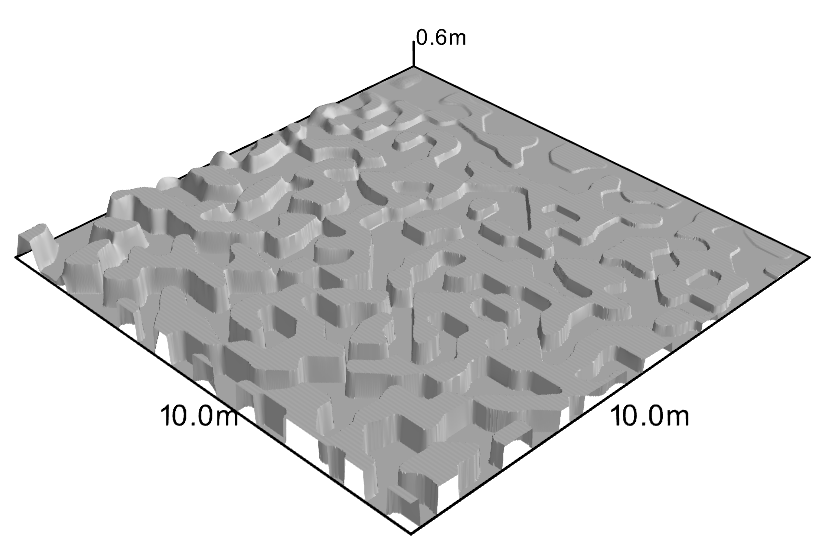}
  \includegraphics[width=0.24\linewidth]{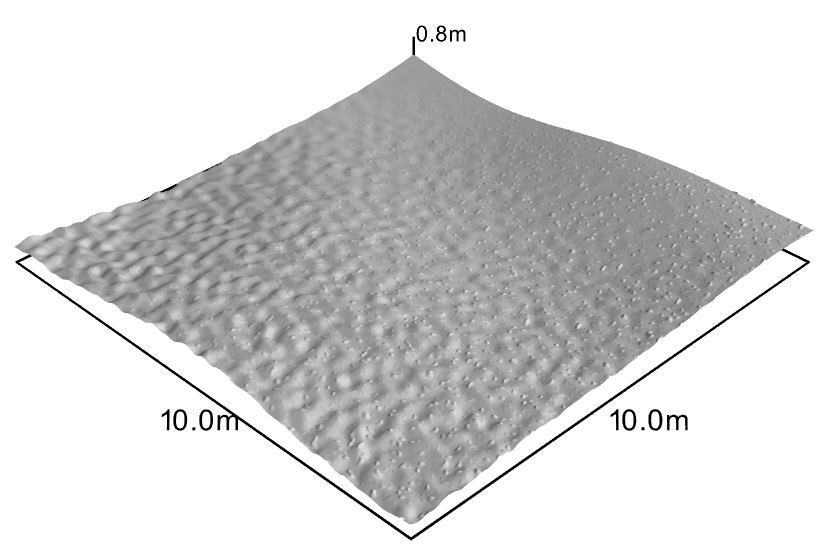}

  \caption{4 out of 30 procedurally generated training terrains, which include common terrain features such as bumps, rails, steps and holes.
  }
  \label{fig:heightmapsexamples}
\end{figure*}

\subsubsection{Simulation Process}

Each simulation begins by picking one at random of the 30 training terrains; the robot is set to a random pose on the map and moves forward at constant velocity without steering. After the robot reaches the edge of the map or gets stuck for some time, it is re-spawned to a different pose on a different training terrain to generate a new trajectory.

For each trajectory, the robot poses and their associated heightmap patches are recorded. The heightmap patch associated to a pose is centred on the robot's position and oriented in such a way that the robot is facing towards the right of the patch (see Fig.~\ref{fig:pipeline}-middle top). If and only if the distance between the current pose $X^\text{robot}(t)$ and a future pose $X^\text{robot}(t+T)$ is greater than a threshold $d$
and aligned with the robot's orientation, then the current patch is labeled as traversable. Otherwise the current patch is labeled as non-traversable.

\begin{figure}
  \centering
  \includegraphics[width=0.99\linewidth]{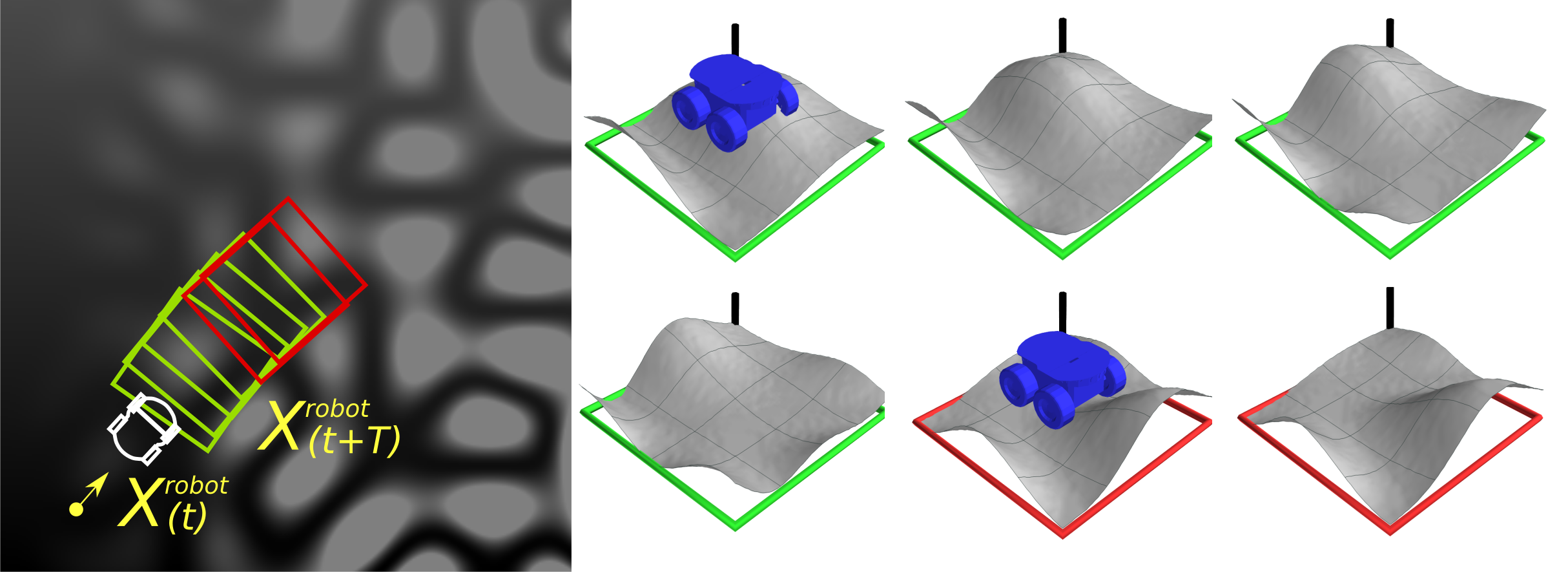}
  \caption{Left: trajectory extracted from a training dataset; robot silhouette and yellow arrow indicate the initial pose.
  Right: visualization of some of the resulting traversable (green) and non-traversable (red) patches.
 }
  \label{fig:trajectoryexamples}
\end{figure}

Figure~\ref{fig:trajectoryexamples}-left illustrates the traversability labeling for patches along a trajectory, with $T=\SI{1}{s}$ and $d=\SI{0.12}{m}$.
The robot traverses a set of smooth patches until it gets blocked by a bump almost as high as the robot itself.
As illustrated in Fig.~\ref{fig:trajectoryexamples}-right, we use this information to label the first patches as traversable and the last patches as non-traversable.

\subsection{Dataset Generation}
\label{sec:datasetgeneration}

From each simulated trajectory (on average \SI{20}{\second}), we sample the robot pose and the corresponding heightmap patch at \SI{20}{\hertz}. The patch and its traversability label represents a sample in the dataset.

We produce a training dataset (450k samples,
equaling roughly \SI{27}{\km} traveled)
using the 30 training terrains and a synthetic evaluation dataset (150k samples) from 10 terrains generated with a similar approach (see Section~\ref{sec:heightmapgeneration}).

Figure~\ref{fig:trajectoryexamples} illustrates the patch extraction process from the trajectory in the synthetic map on the left. The height values in each patch (Fig.~\ref{fig:trajectoryexamples}-right) are offset in such a way that the patch center (i.e. at the robot's position) is mapped to height 0.  This makes the patches independent on their absolute height on the heightmap, a feature that does not affect traversability.

For quantitatively evaluating our classifier in real-world maps, we consider 6 scenarios, from which we extract a total of 350k samples,
equaling roughly \SI{21}{\km} traveled.
Three of these scenarios are publicly-available~\cite{senseflydatasets2016} heightmaps from a mining \emph{quarry}, the town of \emph{Sullens}, and a \emph{Gravel pit}; they have been acquired with a fixed-wing flying robot using structure-from-motion techniques.
One map was built by a quadcopter equipped with a depth sensor, from an indoor set-up in the ETH Autonomous Systems Lab (\emph{ETH-ASL})~\cite{kaslin2016collaborative}.
Additionally, we consider two maps for qualitative assessment of traversability estimation with a real robot:
\emph{Bars}, a simple indoor map consisting of a floor with horizontal bars with different width and height, manually built to verify if the learned model can properly estimate the traversability of a real Pioneer 3-AT;
\emph{Slope}, an outdoors scenario (manually mapped with a Tango RGB-D sensor) consisting of an irregular grass slope surrounded by a smooth uphill walkway. The real robot's performance in this scenario is reported in Section~\ref{sec:pathplanning}.

\begin{figure*}
  \centering
  \includegraphics[width=0.99\linewidth]{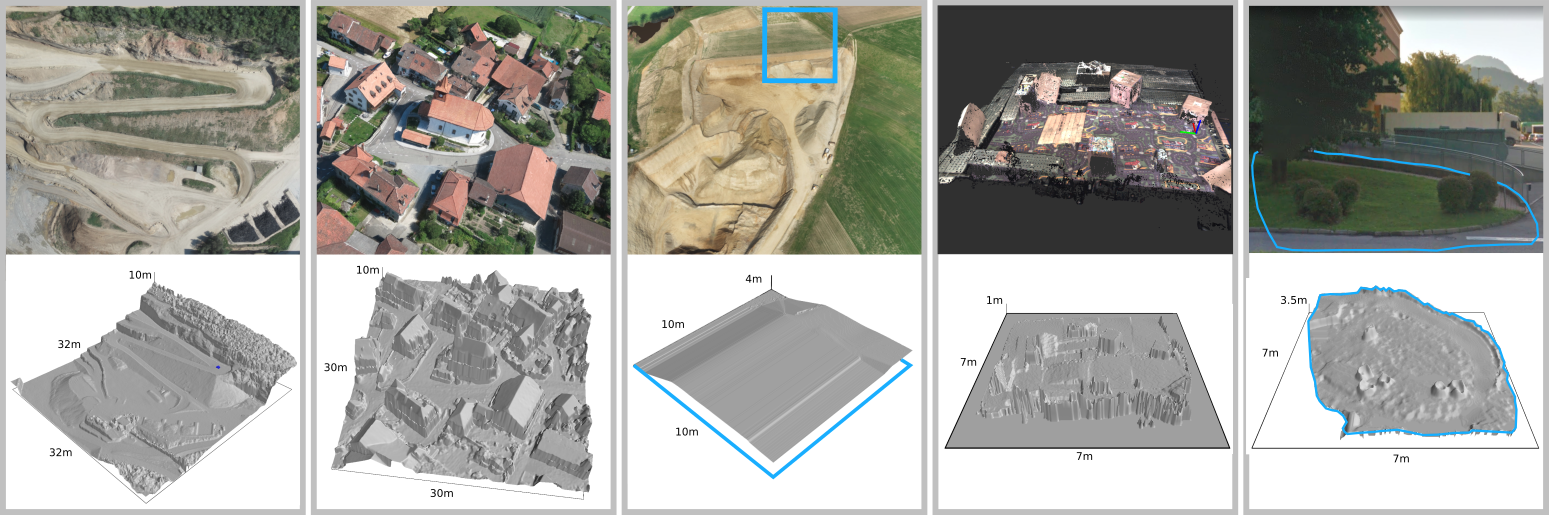}
  \caption{Reference images (top) and heightmaps (bottom) for the 5 of the 6 real-world evaluation datasets.  Left to right: Mining quarry~\cite{senseflydatasets2016}, Sullens~\cite{senseflydatasets2016}, Gravelpit~\cite{senseflydatasets2016} (one of three areas extracted from the same scenario), ETH-ASL\cite{kaslin2016collaborative} and Slope (acquired area corresponds to the blue outline).
  }
  \label{fig:heightmaps_datasets}
\end{figure*}

Figure \ref{fig:heightmaps_datasets} shows reference images of all the real-world datasets (excluding the Bars map) and their corresponding heightmap.
Table~\ref{tab:fullevaluation} summarizes datasets and the maps they have been obtained from.  All datasets were generated using the robot model described in Sec.~\ref{sec:traversabilitysimulation}, $T = \SI{1}{\second}$, $d=\SI{12}{\cm}$ and a patch size equivalent to $\SI{1.2}{m}\times\SI{1.2}{m}$.

\subsection{Training Traversability Classifiers}
\label{sec:traversabilitylearning}

We address the problem of estimating terrain traversability as classification on heightmap data.  We compare two alternative approaches: extracting descriptive features from each heightmap patch and then applying standard statistical classification techniques~\cite{bishop2006}, or adopting CNNs, a now-standard deep-learning approach
which operates directly on raw input data.  In either case, the output of the classifier indicates if a patch is traversable.

For the \textbf{feature-based approach}, we compute quantities that provide traversability cues, such as the average terrain steepness in the robot's motion direction (i.e. from left to right of the patch), or the maximum height of any steps in patch.  In our case, we compute the Histogram of Gradients (HOG) of the heightmap patch, which includes these pieces of information (e.g., the gradient of a heightmap corresponds to the local steepness). Computing HOG over 6 orientations, $\SI{8}{px}\times \SI{8}{px}$ per cell, and a block of $3\times3$ cells, results in a descriptor with 324 features that we classify by means of a Random Forest (RF) classifier~\cite{breiman2001} with 10 trees.

In the \textbf{CNN-based approach}, it is expected that the network autonomously learns meaningful, problem-specific features; because the input data is high-dimensional and no prior knowledge of the problem is provided to the model, this approach requires more training data, which however is available from our extensive simulations.
Our CNN is built on the Keras~\cite{chollet2017deep} front-end powered by TensorFlow~\cite{tensorflow2015}; it is composed by interleaved convolutional and max-pooling layers, a classic architecture that is well known to be suited to visual pattern recognition problems~\cite{ciresan2012multi};
a $\SI{60}{px}\times\SI{60}{px}$ input layer is followed by: a $3\times3$ convolution layer with $5$ output maps; a $3\times3$ convolution layer with $5$ output maps; a $2\times2$ Max-Pooling layer; a $3\times3$ convolution layer with $5$ output maps; a fully connected layer with $128$ output neurons; a fully connected layer with 2 output neurons followed by a softmax layer (output). All layers implement the ReLU activation function.  The network is trained for 100 epochs to minimize a categorical cross-entropy loss using the Adadelta optimizer. Following the current best practices for visual pattern recognition CNNs~\cite{chollet2017deep}, several variations on this architecture have been evaluated with minimal performance differences.

\section{Experimental Results}
\label{sec:experimentalresults}

The following sections detail the comparison of the two classifiers described in Section~\ref{sec:traversabilitylearning} and an additional baseline dummy classifier that always returns the class most frequent in the training set.

\subsection{Classification Results}
\label{sec:classificationresults}

The performance of the three estimators on the evaluation datasets
is reported in Table~\ref{tab:fullevaluation}.
We observe that the CNN estimator outperforms both the baseline and feature-based approaches both on synthetic and real-world heightmaps. Performance is lower on the real datasets Sullens, Gravelpit and Quarry
than on synthetic evaluation data,
probably because of some elevation patterns, such as narrow passages, tight rails and low barriers that may block the robot, which are not well represented in procedurally-generated training data. Although the Sullens dataset
contains a limited variety of interesting terrain features (mainly slopes, steps and bumps), it illustrates that the classifier has a reasonable performance on a common scenario.
Many features of real terrains such as slopes, holes, rails, steps and bumps are correctly classified in real datasets (see Fig. \ref{fig:resultsminingquarry} and \ref{fig:planning}).

In order to validate whether the classifier is effective in a real scenario, we built the Bars environment (Figure~\ref{fig:pipeline}) which features two horizontal wooden bars with a rectangular cross section as obstacles.  One bar is \SI{6}{cm} deep whereas the other is \SI{8}{cm} deep.  Both are \SI{6}{cm} high.  We verified that the robot is unable to traverse the former and can very rarely traverse the latter; the classifier outputs very low traversability probabilities for both; more interestingly, the classifier correctly predicts that the robot would be able to traverse the obstacle if it can climb a bar with only one of its wheels (estimated probability about 50\%); a video demonstration can be found in \materialurl.  A more extensive real-robot experiment in an outdoor scenario is reported in Section~\ref{sec:pathplanning}.

\begin{table*}\centering
  \ra{1.2}
  \cs{3}
  \footnotesize
  \caption{
  For each dataset we report accuracy (ACC) and the area under the ROC curve (AUC) to compare our approach (CNN) to a Feature Based (FB) and a Baseline (BL) classifier.
  }
  \label{tab:fullevaluation}
  \begin{tabular}{@{}llccccccccccccccccccc@{}}
  \toprule
     & & & \multicolumn{10}{c}{Quantitative evaluation in simulation} &&  &&  &&  &&  \\
       \cmidrule{4-13}
     \multicolumn{2}{c}{Dataset} & & &  & \multicolumn{2}{c}{CNN} && \multicolumn{2}{c}{FB} && \multicolumn{2}{c}{BL} && \multirow{2}{*}{\makecell{Real-robot \\tests}} && \multirow{2}{*}{\makecell{Size \\ ( ($\SI{}{m})\times$maps ) }} && \multirow{2}{*}{\makecell{Resolution \\ ($\SI{}{cm/px}$)}} && \multirow{2}{*}{Mapping} \\
      \cmidrule{1-2}  \cmidrule{6-7} \cmidrule{9-10} \cmidrule{12-13}
      Type     &  Name  && Samples  &    &  ACC  &  AUC  &&  ACC  &  AUC  &&  ACC  &  AUC  &&  &&  &&  && \\
  \toprule
\multirow{2}{*}{Synthetic}  & Training   && 450k && - & - && - & - && - & - && - && $(10\times10\times2)\times30$ && 2 && -\\
                           & Evaluation  && 150k && 0.926 & 0.970 && 0.703 & 0.746 && 0.544 & 0.527 && - && $(10\times10\times2)\times10$ && 2 && - \\
\cmidrule{2-21}
\multirow{6}{*}{\makecell[l]{Real\\evaluation}}   & Quarry    && 100k  && 0.819 & 0.840 && 0.723 & 0.762 && 0.542 & 0.499 && - && $32\times32\times10$ && 2 && Fixed-wing \\
                         & Sullens  && 100k  && 0.858 & 0.884 && 0.781 & 0.799 && 0.527 & 0.501 && - && $30\times30\times10$ && 3 && Fixed-wing  \\
                         & Gravelpit && 60k  && 0.832 & 0.839 && 0.769 & 0.802 && 0.577 & 0.504 && - && $(10\times20\times4)\times3$ && 5 && Fixed-wing  \\
                         & ETH-ASL  &&  30k  && 0.887 & 0.921 && 0.789 & 0.817 && 0.492 & 0.511 && - && $7\times7\times1$ && 2 && Quadcopter \\
                         & Slope    &&  40k  && 0.864 & 0.875 && 0.761 & 0.780 && 0.530 & 0.501 && Yes && $7\times7\times2.5$ && 2 && Handheld \\
                         & Bars     &&  20k  && 0.961 & 0.980 && 0.829 & 0.863 && 0.493 & 0.498 && Yes && $10\times10\times1$ && 5 && CAD \\
  \bottomrule
  \end{tabular}
  \normalsize
\end{table*}

\subsection{Traversability Visualization on the Quarry Dataset}
\label{sec:traversability_estimation_real}

The Quarry dataset was generated from a mining quarry map (see Fig. \ref{fig:heightmaps_datasets}) of \SI{0.02}{\square \km}~\cite{senseflydatasets2016}. This map contains challenging roads and barriers designed for mining trucks.
To make the map more suitable for our robot,
we re-scaled the map to $\frac{1}{4.5}$ of its original size.

\begin{figure}
  \centering
  \includegraphics[width=0.9\linewidth]{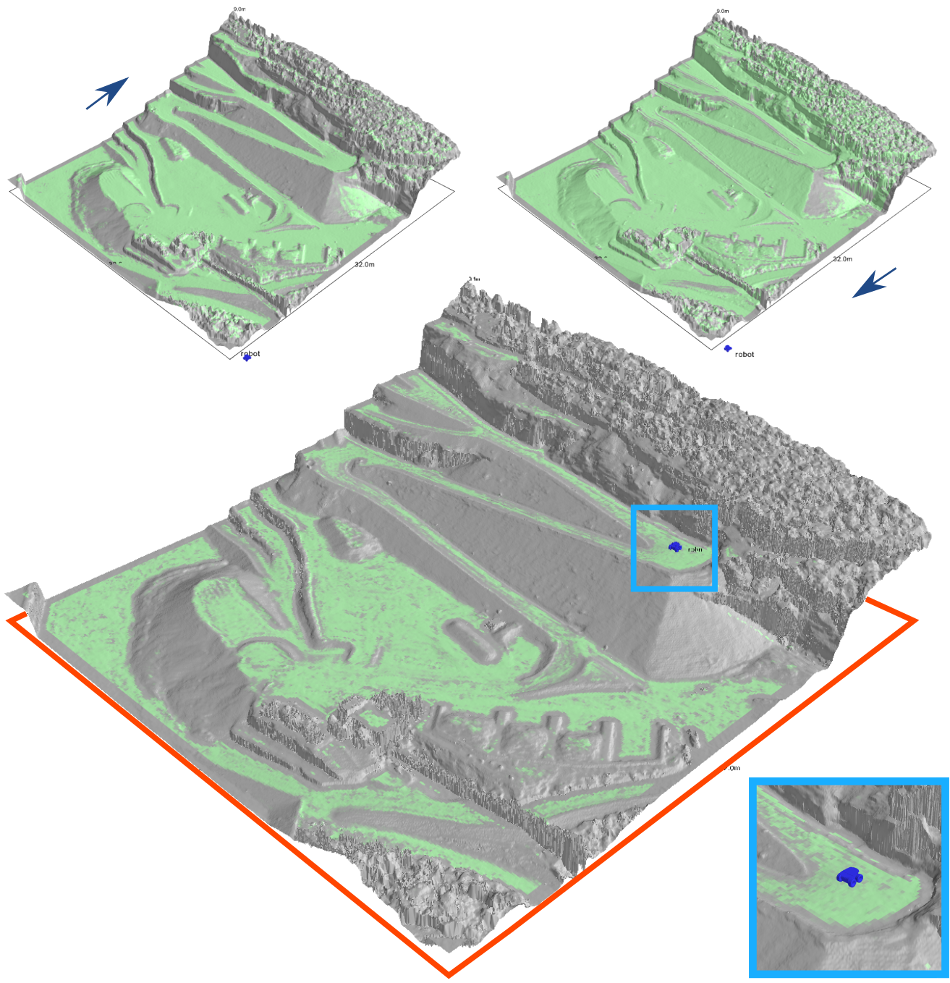}
  \caption{Top: Traversability estimation for the Quarry map for two orientations: 
  $\ang{90}$ (up) and $\ang{270}$ (down).
  Bottom: Minimum traversability (orange frame) overlay over 32 possible orientations.
  Green opacity indicates traversability degree.
  Pioneer 3-AT's model is displayed for size comparison.
  }
  \label{fig:resultsminingquarry}
\end{figure}

For this analysis, we fix a direction and iterate over the entire heightmap extracting patches of $\SI{60}{px}\times\SI{60}{px}$ with a stride of \SI{5}{px}. This process is equivalent to translating the robot's position over the map while keeping a fixed orientation.
Figure~\ref{fig:resultsminingquarry}-top shows the traversability estimation for the mining quarry for two orientations, indicated by the arrows.
Traversability is represented as a colored overlay on the surface of the heightmap (traversable is green, non-traversable is gray). 

The estimator correctly marks the main road ($\approx\SI{2}{m}$ wide) as traversable.
Very steep or vertical slopes are recognized as non-traversable when going up but, dangerously, marked traversable when going down. This particular result is congruent with the definition of traversability we used, which only considers whether the robot can proceed but does not account for its safety.
Slopes are always classified as traversable downhill and, sometimes, transversally.
Most of the rough surface
at the top of the heightmap is correctly found as non-traversable.
The orange frame in Fig.~\ref{fig:resultsminingquarry} reports the minimum traversability over 32 orientations: green areas are traversable in all directions; this yields a compact, intuitive visualization but does not convey directional information.

\begin{figure}
  \centering
  \includegraphics[width=0.93\linewidth]{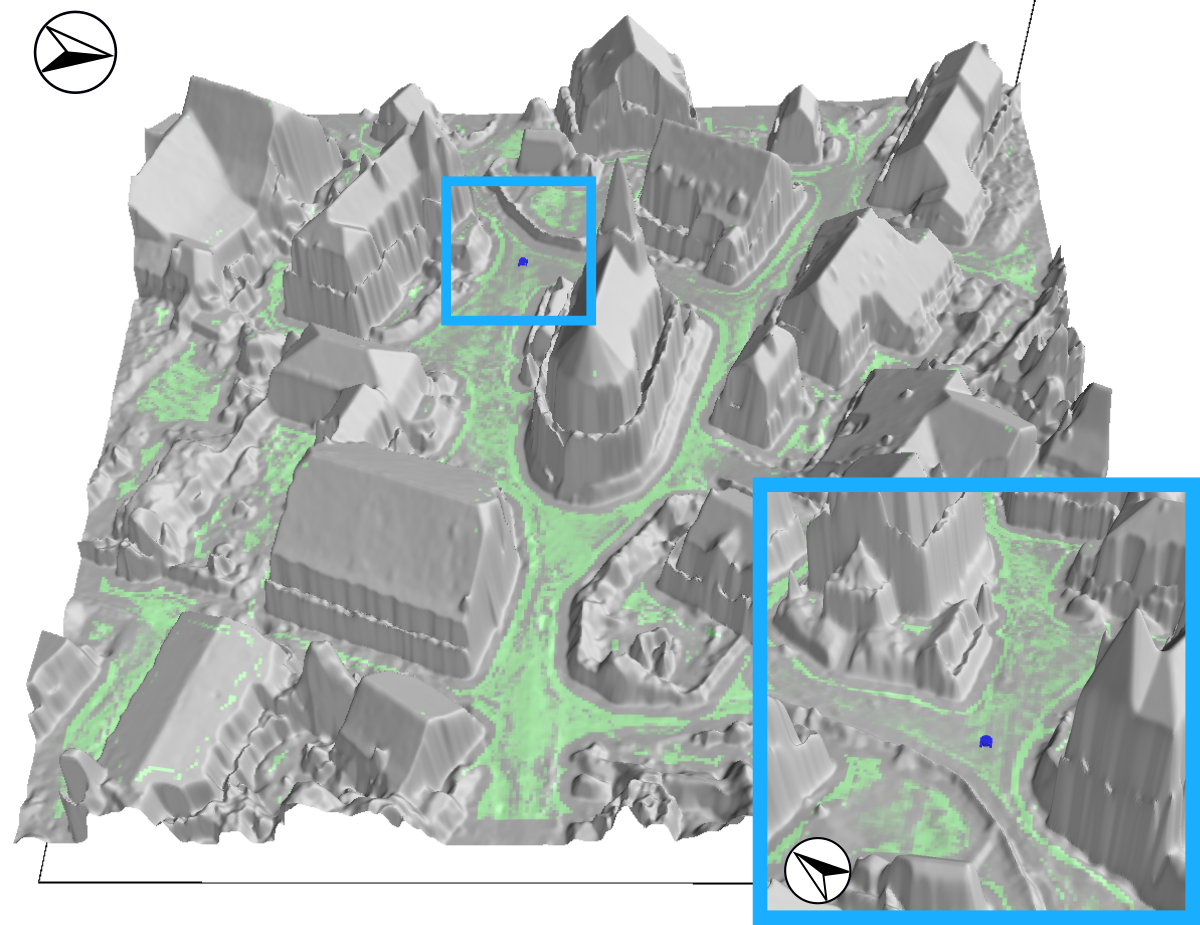}
  \caption{Minimum traversability overlay of the Sullens map over 32 possible orientations. Blue square zooms in a region facing the central building.}
  \label{fig:sullens}
\end{figure}

Traversable areas in the mining quarry correspond to mining roads, flat terrain, short bumps and slopes. Figure~\ref{fig:sullens} shows the minimal traversability map for the Sullens heightmap. Streets, pedestrian side-walks and garden entries are clearly estimated as traversable. Challenging slopes uphill to the central building (blue square) are classified with an average confidence of traversability (translucent green).

\subsection{Computational cost}

Training the traversability classifier requires to: procedurally generate terrains (few milliseconds); simulate the robot on such terrains to generate the 450k-sample training datasets (about one day); actually train the CNN (about 1 hour on a single NVidia Tesla K80 GPU).

Once the classifier is trained, building traversability maps requires to evaluate a large amount of patches; considering a stride of \SI{10}{cm} and 32 orientations, we need to evaluate $10 \times 10 \times 32 = 3200$ patches per square meter, which takes \SI{350}{\ms} on an Intel i7 desktop computer on CPU, and could be easily parallelized to run on GPU; this timescale allows to densely process maps at a rate comparable to the time needed to acquire them.  The time needed for inference is independent on the content of the patches.

The computational expense for simulating a robot, in contrast, depends on the complexity of the heightmap; in our setup, which is not optimized for speed, simulation speed (excluding visualization) ranged from 0.1$\times$ real time for very complex maps (e.g., quarry) to 10$\times$ real time for flat areas, averaging at 1$\times$ real time on relatively easy terrain. Simulating the traversal of a \SI{1}{\square\meter} area could be approximated by simulating 32, \SI{1}{\meter} long, trajectories, one for each orientation; at a robot speed of 10 cm per second and 1x real time factor, this requires 320 seconds, i.e., almost $1000\times$ the time required by our approach.  These numbers may drastically change if consider the extra cost for more sophisticated simulation (e.g., for soft ground), or the speedup granted by future GPU-optimized simulators.

\subsection{Path Planning on Probabilistic Traversability Maps}
\label{sec:pathplanning}

Traversability estimations can be used to plan a path to a goal position which only passes through patches that are traversable in the direction corresponding to the path's local orientation.
In the following, we assume that the robot will get stuck (or crash, or end up in an unrecoverable state) if it attempts to pass through a non-traversable patch.\footnote{This is a pessimistic assumption because this is not always the case: a robot attempting to pass through a wall, for example, is unable to proceed but may be capable of turning back and taking another path.}

We assume that traversability probabilities of patches along a path are independent from each other, i.e., that the traversability along a path is Markovian.
If we further assume that the robot is able to rotate in place, the probability $\pi$ to traverse a path composed of segments $(e_1, \ldots, e_n)$ is given by
$\pi((e_1, \ldots, e_n)) = \prod_{i=1}^n \pi(e_i)$, where $\pi(e_i)$ is the probability to traverse a segment.

However, on unstructured terrains, the robot may not be able to rotate in place everywhere.  Following the same pipeline introduced for traversability estimation, we train a \emph{turnability} classifier to estimate the probability that the robot, positioned in the center of a given patch, can rotate in place by at least \ang{45}, clock or counter-clock wise. The performance of these classifiers on the Synthetic and Quarry evaluation datasets was (ACC 0.843, AUC 0.874) and (ACC 0.828, AUC 0.834) respectively.

\begin{figure}
  \centering
 \includegraphics[width=0.99\linewidth]{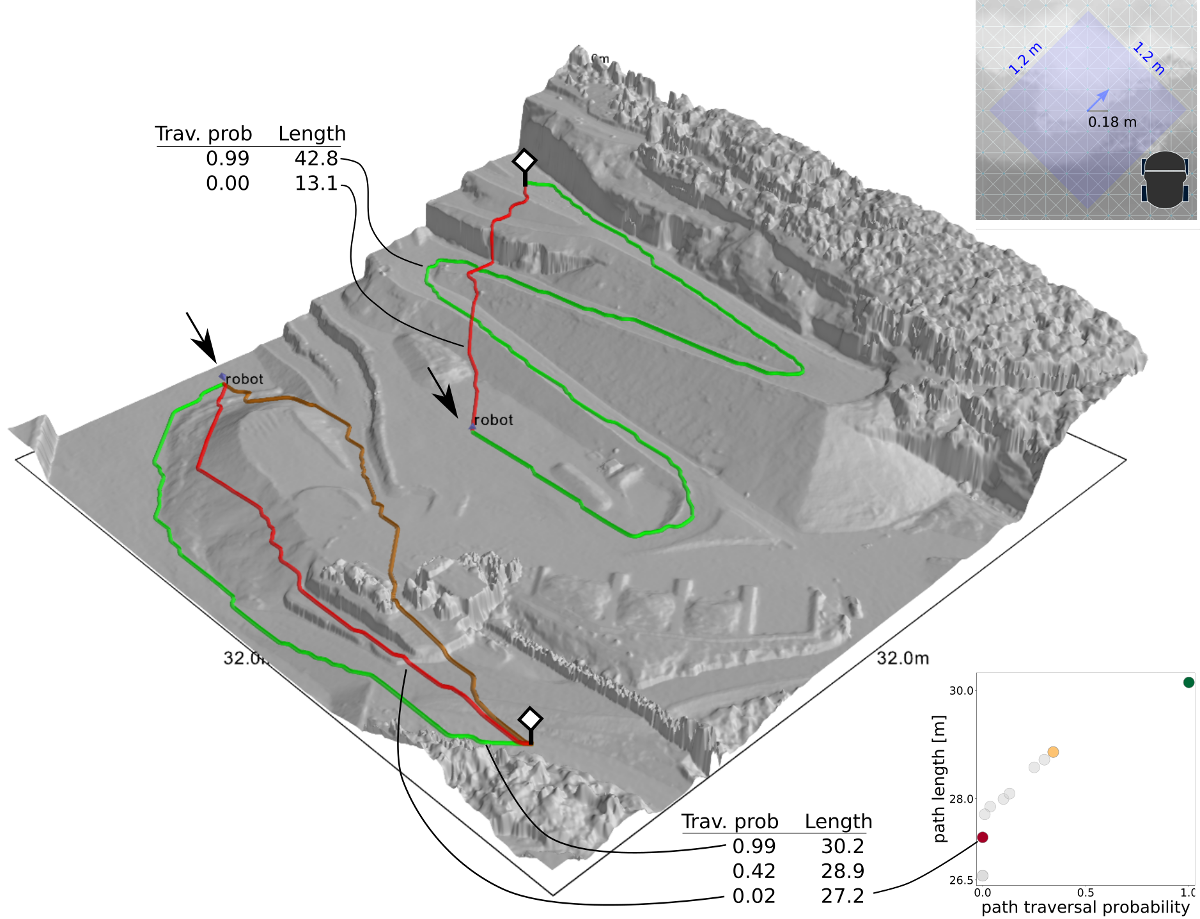}
  \caption{Left: a selection of Pareto optimal paths on the quarry map for two pairs of source (arrow) and target (white square) locations. The paths are colored by traversal probability from red (non-traversable) to green (surely traversable). Corresponding values for traversal probability and length of each trajectory are shown in the side tables.
Right top: a portion of the planning graph on the quarry map. Nodes are placed on a regular grid at \SI{18}{cm}. The blue edge's traversal probability is estimated by applying the classifier on a $\SI{1.2}{m} \times \SI{1.2}{m}$ patch (light blue) centered at the edge origin and directed along the edge. Robot's silhouette is shown for size comparison. Right bottom: The trade-off between path length and traversability for Pareto optimal paths of the bottom source-target location. Colored dots correspond to paths drawn on the left figure.}
  \label{fig:planning}
\end{figure}

We compute paths on a graph $(N, E)$ of nodes regularly distributed on a horizontal grid of poses of spatial side \SI{18}{cm} and angular side \ang{45}. Edges in $E$ comprise  rotations in place by $\pm \ang{45}$ and segments connecting poses of the same orientation that are spatial neighbors.
For each edge $e = (n,m) \in E$,  we compute length and traversal probability $\pi(e)$. $\pi(e)$ is the output of the traversability or turnability classifiers applied to a patch centered and oriented along $n$ (see Fig.~\ref{fig:planning}-right top).

Figure~\ref{fig:planning}-right bottom illustrates the solution of the multi-objective problem of computing the best paths with respect to length (minimized) and traversability (maximized). A rational agent would choose among the set of Pareto-optimal paths.  In this context, a path is Pareto-optimal if there exists no alternative path that is both more traversable and shorter.  The shortest path (represented in red in Fig.~\ref{fig:planning}-left) is Pareto-optimal, but will have in most cases a very low traversability probability.  The path with the maximum traversability (represented in green) will also be Pareto-optimal, but may be unnecessarily long; between the two extremes, a potentially very large set of Pareto-optimal paths exist, spanning the trade-off between traversability and length.

\paragraph*{Real-robot experiment}

We tested how our approach applies to a real Pioneer 3-AT robot in an outdoor grass slope (Slope map in Fig.\ref{fig:heightmaps_datasets}). First we use a Tango device, handheld at \SI{1}{m} from the ground pointing downwards, to build a heightmap map of the area. Then we apply our approach to compute the oriented traversability maps and the traversability graph. 
We then evaluate the safest (i.e. with maximal traversability) path from a source position (blue silhouette of the robot) to a desired position in the map (white marks).
The safest way for the robot to reach the top area (square mark) is to follow the smooth side-walk ramp uphill, avoiding the grass slope which has an irregular shaped terrain. This path is estimated as certainly traversable: we verified this is in fact a traversable path by teleoperating the robot through it.
The maximal-traversability path that reaches the circle mark involves first reaching the square mark uphill on the sidewalk, then heading down on the grassy slope for a short distance: even though it is long, this path is in fact the most rational to reach such point, because traversing the grass slope uphill or transversally is challenging.
The star mark lies on a difficult-to reach area in the middle of the grassy slope.  The point is not reachable from the sidewalk above it, because such area of the sidewalk is flanked by small step that is correctly estimated as not traversable.  The maximal-traversability path, instead, accesses the grass in a point left to the robot, then proceeds uphill avoiding obstacles and excessively steep or rugged areas; the path has a traversability probability of 0.21, and we were unable to successfully teleoperate the robot through it because it was blocked by a bump.

\begin{figure}
  \centering
 \includegraphics[width=0.97\linewidth]{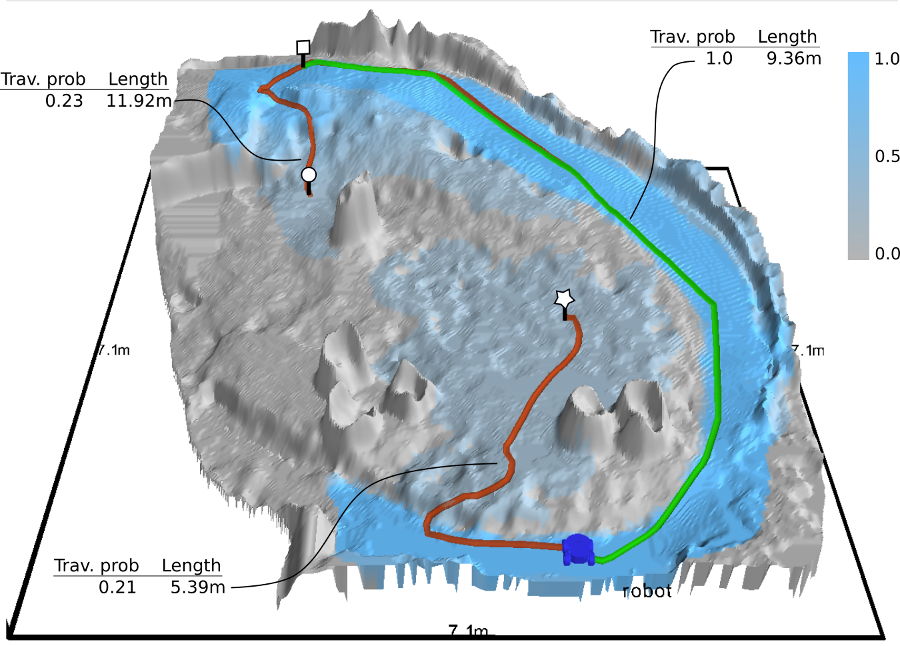}
\caption{
Paths of maximal traversability from the robot's initial pose (blue silhouette) to three different goals in the Slope map. Paths are colored according to their traversal probability, from red (low) to green (high).
The blue overlay represents the \emph{reachability} map from the robot's initial pose: blue (certainly reachable) to gray (certainly not reachable).
}
  \label{fig:pioneer}
\end{figure}

The \emph{reachability} map in Fig.\ref{fig:pioneer} illustrates which parts of the terrain the robot can \emph{reach} from its current pose. For a given target location, it is defined by the maximal traversal probability among all paths from source to target.

\section{Discussion and Conclusions}
\label{sec:discussion}

We presented a complete framework for traversability estimation that casts the problem as a heightmap classification task, and applies to any modality of robot locomotion.  Classifiers trained on simulation data using procedurally-generated terrains capture relevant terrain characteristics and can efficiently and accurately estimate oriented traversability maps on large unseen real-world terrains. Such maps can be used to plan paths exploring the trade-off between length (to be minimized) and probability of being traversable (to be maximized).

Our approach does not capture the robot dynamics, such as the speed with which the robot approaches an obstacle: we limit our attention to slow robots operating on rugged terrains, where dynamic aspects have negligible impact. We will consider dynamic data as an additional input of our classifier as we move towards scenarios where complex locomotion is needed.

We remark two important limitations of our approach. 1) We assume that the terrain's 3D shape is the only factor influencing its traversability; this is an acceptable approximation in some scenarios (e.g., on solid ground where friction is not a key factor), but is inadequate in many others, such as on sand, mud, wet or slippery environments. 2) We only use simulated data for training, but in practice, simulations rarely manage to accurately replicate the real world.

In order to tackle both limitations, future work will aim at using real data for training (as in~\cite{karumanchi2010}), which fits well our conceptual framework. Technically, this requires that the robot can associate the heightmap patches it's traversing with an outcome (successfully traversed or not) in order to accumulate experience that can be used for training new models or refining existing ones.  Note that, in this scenario, the robot would automatically learn the effects of soft ground or slipping on traversability.

A notable exception are scenarios in which 3D shape alone is insufficient to predict traversability: for example, a gentle paved slope could be traversable, unlike a gentle muddy slope which has the same shape. In our framework, one could differentiate these cases by learning different models for different terrain types (to be previously classified based on their appearance as in~\cite{delmerico2016}) or by training a single model that uses terrain appearance as an additional input in addition to 3D shape.

\bibliographystyle{IEEEtran}
\bibliography{library}

\end{document}